\documentclass[twoside,11pt]{article} 
\usepackage{jmlr2e} 

\jmlrheading{}{}{}{}{}{Marc Claesen, Frank De Smet, Pieter Gillard, Chantal Mathieu and Bart De Moor}

\ShortHeadings{Predicting the Start of Glucose-Lowering Pharmacotherapy}{Marc Claesen, Frank De Smet, Pieter Gillard, Chantal Mathieu and Bart De Moor}
\firstpageno{1}

\usepackage{amsmath}
\usepackage{tikz}
\usepackage{ifpdf}
\usepackage{multirow}
\usepackage{afterpage}
\usepackage{booktabs}
\usepackage{fancyvrb}
\usepackage{color}
\usepackage[colorlinks=false,allbordercolors={1 1 1}]{hyperref}
\usepackage{listings}
\usepackage{xspace}
\usepackage{algorithmicx}
\usepackage{subfigure} 

\newcommand{\dmeds}{d_{\rm meds}}
\newcommand{\dprovs}{d_{\rm provs}}

\begin{document}

\title{Building Classifiers to Predict the Start of Glucose-Lowering Pharmacotherapy Using Belgian Health Expenditure Data}

\author{\name Marc Claesen \email marc.claesen@esat.kuleuven.be  \\ 
\addr KU Leuven, Department of Electrical Engineering (ESAT) \\
STADIUS Center for Dynamical Systems, Signal Processing and Data Analytics \\ 
iMinds, Department of Medical Information Technologies \\
Kasteelpark Arenberg 10 - box 2446, 3001 Leuven, Belgium \\
\name Frank De Smet \email frank.desmet@cm.be\\ 
\addr National Alliance of Christian Mutualities \\ 
Haachtsesteenweg 579, 1031 Schaarbeek, Belgium \\
\addr KU Leuven, Department of Public Health and Primary Care, Environment and Health \\
Kapucijnenvoer 35 block d - box 7001, 3000 Leuven, Belgium 
 \AND
\name Pieter Gillard \email pieter.gillard@uzleuven.be \\
\name Chantal Mathieu \email chantal.mathieu@uzleuven.be \\
\addr KU Leuven, Department of Clinical and Experimental Endocrinology \\
UZ Herestraat 49 - box 902, 3001 Leuven, Belgium 
 \AND
\name Bart De Moor \email bart.demoor@esat.kuleuven.be  \\ 
\addr KU Leuven, Department of Electrical Engineering (ESAT) \\
STADIUS Center for Dynamical Systems, Signal Processing and Data Analytics \\ 
iMinds, Department of Medical Information Technologies \\
Kasteelpark Arenberg 10 - box 2446, 3001 Leuven, Belgium
}

\editor{TBD}

\maketitle

\begin{abstract}
Early diagnosis is important for type 2 diabetes (T2D) to improve patient prognosis, prevent complications and reduce long-term treatment costs. We present a novel risk profiling approach based exclusively on health expenditure data that is available to Belgian mutual health insurers. We used expenditure data related to drug purchases and medical provisions to construct models that predict whether a patient will start glucose-lowering pharmacotherapy in the coming years, based on that patient's recent medical expenditure history. The design and implementation of the modeling strategy are discussed in detail and several learning methods are benchmarked for our application. Our best performing model obtains between $74.9\%$ and $76.8\%$ area under the ROC curve, which is comparable to state-of-the-art risk prediction approaches for T2D based on questionnaires. In contrast to other methods, our approach can be implemented on a population-wide scale at virtually no extra operational cost. Possibly, our approach can be further improved by additional information about some risk factors of T2D that is unavailable in health expenditure data.
\end{abstract}

\begin{keywords}
diabetes, public health, insurance data, classification, risk analysis
\end{keywords}

%
%

\section{Introduction}
Type 2 diabetes mellitus (T2D) is a chronic metabolic disorder characterized by hyperglycemia and is considered one of the main threats to human health \citep{zimmet2001global}. In developed countries, T2D makes up about 85\% of diabetes mellitus patients and occurs when either insufficient insulin is produced, the body becomes resistant to insulin or both \citep{world1994prevention}. Prediabetes and less severe cases of T2D are initially managed by lifestyle changes, specifically increasing physical exercise, dietary change and smoking cessation \citep{tuomilehto2001prevention, diabetes2002reduction, american2014standards}. If this yields insufficient glycemic control, pharmacotherapy with glucose-lowering agents (GLAs) like metformin or insulin is started \citep{turner1999glycemic, american2014standards}.

Several studies have indicated that one third to one half of T2D patients are undiagnosed \citep{harris1998prevalence, king1998global, rubin1994health}. Additionally, patients often remain undiagnosed for extended periods of time, with average diagnose-free intervals ranging from 4 to 7 years \citep{harris1992onset}. The prognosis of untreated patients can deteriorate rapidly as prolonged hyperglycemia can cause serious damage to many of the body's systems. Timely diagnosis of T2D proves challenging in contemporary medicine, as many patients already present signs of complications of the disease at the time of clinical diagnosis of T2D \citep{harris1998risk, rajala1998prevalence, kohner1998united,ballard1988epidemiology,harris2000early, hu2002elevated}.

Earlier diagnosis and subsequent treatment is believed to prevent or delay complications and improve prognosis \citep{pauker1993deciding, engelgau2000screening}. When impaired glucose tolerance is diagnosed early, initial treatment can often be limited to lifestyle changes \citep{pan1997effects,tuomilehto2001prevention,diabetes2002reduction}. Compared to pharmacotherapy, lifestyle changes are simple, fully manageable by the patient and far less likely to cause serious treatment-induced complications like hypoglycemia \citep{seltzer1989drug, zammitt2005hypoglycemia}. Complementary to health benefits, early diagnosis of T2D poses a health economical advantage, as patients that do not require acute or intensive long-term treatment are far less demanding on the health care system.

Universal screening for T2D is cost-prohibitive \citep{wareham2001should, engelgau2000screening}, but many organizations advise opportunistic screening of high-risk subgroups \citep{world1994prevention, alberti1998report, engelgau2000screening,american2014standards}. Several risk profiling strategies have been developed to aid in the timely diagnosis of T2D \citep{baan1999performance,stern2002identification, lindstrom2003diabetes,mcneely2003comparison,charlone2004danish,heikes2008diabetes,schwarz2009finnish}. Risk profiling is typically done by assessing some of the key risk factors for T2D, which include obesity \citep{mokdad2003prevalence}, genetic predisposal \citep{shai2006ethnicity,interact2013link}, lifestyle \citep{reis2011lifestyle} and various clinical parameters. Existing risk profiling approaches are implemented via questionnaires, potentially augmented with clinical information that is available to the patient's general practitionner \citep{griffin2000diabetes,spijkerman2004performance,lindstrom2003diabetes,glumer2004danish, schulze2007accurate, heikes2008diabetes}. Commonly required information includes BMI, family history, exercise and smoking habits and various clinical parameters.

In this work, we present an alternative approach for risk profiling which only requires data that is already available to Belgian mutual health insurers. This work was done in  collaboration with the National Alliance of Christian Mutualities (NACM). NACM is the largest Belgian mutual health insurer with over four million members. Our approach does not require any questionnaires or additional clinical information and predicts whether a patient will start taking GLAs in the next few years. Interestingly, our approach works well despite the fact that Belgian health insurer data contains little direct information regarding key risk factors of T2D, that is weight, lifestyle and family history are all unavailable.

\section{Existing type 2 diabetes risk profiling approaches} \label{sec:stateoftheart}
The Cambridge Risk Score (CRS) was developed to assess the probability of undiagnosed T2D based on data that is routinely available in primary care records, including age, sex, medication use, family history of diabetes, BMI and smoking status \citep{griffin2000diabetes}, The CRS has been shown to be useful on multiple occasions \citep{griffin2000diabetes, park2002performance, spijkerman2004performance}, though its AUC seems to depend heavily on the population in which it is used, ranging between $67\%$ \citep{spijkerman2004performance} and $80\%$ \citep{griffin2000diabetes}. The information used in the CRS is comparable to another approach which obtained AUCs ranging between $70\%$ and $78\%$ \citep{baan1999performance}.

The FINDRISC score is based on a 10-year follow-up using age, BMI, waist circumference, history of antihypertensive drugs and high blood glucose, physical activity and diet with reported AUCs of $85\%$ and $87\%$ in predicting drug-treated diabetes \citep{lindstrom2003diabetes}. The strongest reported predictors in this study were BMI, waist circumference, history of high blood glucose and physical activity. \citet{glumer2004danish} developed a risk score based on age, sex, BMI, known hypertension, physical activity and family history of diabetes with AUC ranging from $72\%$ to $87.6\%$. The German diabetes risk score reached AUCs ranging from $75\%$ to $83\%$ on validation data and is based on age, waist circumference, height, history of hypertension, physical activity, smoking, and diet \citep{schulze2007accurate}.

\citet{heikes2008diabetes} developed a decision tree for risk prediction achieving $82\%$ AUC in a cross-validation setting, based on weight, age, family history and various clinical parameters. Various other approaches based on routine clinical information have demonstrated similarly accurate predictions of type 2 diabetes \citep{stern2002identification, mcneely2003comparison}.

%
%

\section{Health expenditure data}
The Belgian health care insurance is a broad solidarity-based form of social insurance. Mutual health insurers such as NACM are the legally-appointed bodies for managing and providing the Belgian compulsory health care and disability insurance, among other things. To implement their operations, Belgian mutual health insurers dispose of large databases containing health expenditure records of all their respective members. 

These expenditure records hold all financial reimbursements of drugs, procedures and contacts with health care professionals. Each record comprises a timestamp, financial details and a description of the claim. The financial aspect is irrelevant from a medical point of view, but the type of resource-use as indicated by the description can contain medical information about the patient. These types belong to one of two main categories:
\begin{enumerate}
\item \textbf{Drug purchases} are recorded per package. The coding of packages contains information about the active substances in the drug along with the volume of the package.
\item \textbf{Medical provisions} are identified by a national encoding along with an identifier of the associated medical caregiver. Each provision has a distinct code number.
\end{enumerate}

In addition to resource-use data, some biographical information is available about each patient including age, gender, place of residence and social parameters. In the remainder of this Section we will elaborate on expenditure records related to drugs and provisions. Subsequently we will briefly summarize the main strengths and limitations of using health expenditure data for predictive modeling.

\subsection{Records related to drug purchases}
Expenditure records concerning drug purchases contain information about the active substances in the drug and the purchased volume. We mapped all active substances onto the anatomical therapeutic chemical (ATC) classification system maintained by the \citet{world1996guidelines}. The ATC classification system divides active substances into different groups based on the organ or system on which they act and their therapeutic, pharmacological and chemical properties. Each drug is classified in groups at 5 levels in the ATC hierarchy: fourteen main groups (1st level), pharmacological/therapeutic subgroups (2nd level), chemical subgroups (3rd and 4th level) and the chemical substance (5th level).

After mapping records onto the ATC classification system, a patient's medication history consists of specific ATC codes (5th level) along with the associated number of defined daily doses (DDD). In the period of interest, purchases of 4,580 distinct active substances were recorded in the NACM database. Table~\ref{table:atc-example} shows an example of the classification of active substance on all levels in the ATC system.

\begin{table}[!h]
\centering
\begin{tabular}{cll}
\toprule
level	 & ATC code	 & description \\
\midrule
1 	& A	& alimentary tract and metabolism \\
2	& A10	& drugs used in diabetes \\
3	& A10B	& blood glucose lowering drugs, excluding insulins \\
4	& A10BA & biguanides \\
5	& A10BA02 & metformin \\
\bottomrule
\end{tabular}
\caption{Example of the ATC classification system: classification of metformin per level.}
\label{table:atc-example}
\end{table}

\subsection{Records related to medical provisions}
Expenditure records concerning medical provisions can be considered tuples containing time-stamped identifiers of the patient, physician and medical provision. A single patient-physician interaction may yield multiple such records, one for each specific provision that occurred.

In the Belgian health care system, medical provisions are encoded via the Belgian nomenclature of medical provisions \citep{van2008financing}, which is maintained by the National Institute for Health and Disability Insurance (NIHDI).\footnote{The website of NIHDI is available at \url{http://www.riziv.fgov.be}.} This nomenclature is an unstructured list of unique codes (numbers) for each provision that is being refunded. Nomenclature numbers are added when new provisions are defined or when revisions are made. A single provision may correspond to multiple numbers for various reasons.

\subsection{Advantages of health expenditure data}
The key benefit of expenditure databases is that they centralize structured medical information across all medical stakeholders to yield a comprehensive, longitudinal overview of each patient's medical history. Other health data sources are fragmented, e.g. medical records maintained by the patient's general practitioner or hospital often contain only a subset of the patient's medical history. This fragmentation hampers the identification of patterns that may indicate elevated risk for diseases like type 2 diabetes. The NACM database comprises claims records of over four million Belgians, which enables complex modeling. Additionally, claims data have few omissions due to the financial incentive for patients and medical stakeholders (hospitals) to claim refunds. While other health data sources may contain more detailed information, the strength of NACM's data is in its volume, both in terms of number of patients and the amount of information that is recorded per individual. Finally, as most people tend to stay affiliated with the same mutual health insurer, their expenditure records provide long-term information.

\subsection{Limitations of health expenditure data}
Belgian health expenditure data is strictly limited to what is required for mutual health insurers to implement their operations, which are mainly administrative in nature. Detailed health information such as diagnoses and test results are not directly available. In some other countries, health insurers dispose of more detailed information, such as ICD-10 codes which include diagnoses and symptoms \citep{world2012international}. Including such information is out of scope of this work as we focus exclusively on data that is already available to Belgian mutual health insurers. Biographical information about patients does not contain direct information about some important risk factors such as lifestyle, family history and BMI, though this may be partially embedded indirectly in medical resource-use. 

%
%

\section{Methods}
In this Section we define the prediction task and describe all its aspects: the overall setup (Section~\ref{setup}), the data and its representation (Section~\ref{data}) and the learning algorithms (Section~\ref{learning-methods}). Briefly, our aim is to predict which patients will start glucose-lowering pharmacotherapy within the next 4 years, based on expenditure records of the previous 4 years. 

Our key hypothesis is that patients with increased risk for T2D or those that are already afflicted but not diagnosed have a different medical expenditure history than patients without impaired glycemic control. We essentially use the start of GLA therapy as a proxy for diagnosis of (advanced) type 2 diabetes. This is reasonable since most patients that start GLA therapy above 40 years old have T2D \citep{world1994prevention}.

We posed this task as a binary classification problem. Our classifiers produce a numeric level of confidence that a given patient will start glucose-lowering pharmacotherapy. When predicting a population, the outputs can be used to rank patients according to decreasing confidence that the patients will start glucose-lowering therapy. Highly ranked patients represent a high-risk subgroup which can be targetted for clinical screening.

The full learning setup is described in Section~\ref{setup}, involving different learning methods and representations of patients' expenditure data. Briefly, we used nested cross-validation to obtain unbiased estimates of the predictive performance of each vectorization and learning approach. Predictive performance of all models was quantified via (area under) receiver operating characteristic (ROC) curves. 

\paragraph{Data} Our work is based on a subset of the expenditure records of NACM. All data extractions and analyses were performed at the Medical Management Department of the NACM under supervision of the Chief Medical Officer. The other research partners received no personally identifiable information (including small cells) from NACM. The patient selection and vector representations are described in detail in Section~\ref{data}.

\paragraph{Class definitions}
The positive class was defined as patients that require GLAs for long-term glycemic control.\footnote{GLAs are defined as any drug in ATC category \texttt{A10}, which includes metformin, sulfonylurea and insulin.} The negative class is then defined as patients that do not need GLAs. Expenditure records related to GLAs were used to identify a set of known positives. However, the absence of such records in a patient's resource use history is not proof that this patient has no need for GLAs. This subtle difference is crucial, because it is well known that patients with impaired glycemic control or T2D often remain undiagnosed and hence untreated for a very long time \citep{harris1998prevalence, king1998global, american2014standards}. As we cannot identify negatives, we had to build models from positive and unlabeled data.

\paragraph{PU learning}
Learning binary classifiers from positive and unlabeled data (PU learning) is a well-studied branch of semi-supervised learning \citep{Lee03learningwith,Elkan:2008:LCO:1401890.1401920,MORDELET-2010-523336,Claesen2015resvm}. PU learning is more challenging than fully supervised binary classification, since it requires special learning approaches and quality metrics for hyperparameter optimization that account for the lack of known negatives. We benchmarked three PU learning methods, which are discussed in more detail in Section~\ref{learning-methods}. 

\paragraph{Software}
The entire data analysis pipeline was implemented using open-source software. For general data transformations and preprocessing we used \emph{SciPy} and \emph{NumPy} \citep{scipy,van2011numpy}. The learning algorithms we used are available in \emph{scikit-learn} and \emph{EnsembleSVM} \citep{pedregosa2011scikit, JMLR:v15:claesen14a} . Finally, we used \emph{Optunity} for automated hyperparameter optimization \citep{claesen2014easy}.


\subsection{Experimental setup} \label{setup}
We gathered all expenditure records during the 4-year interval of 2008 up to 2012. The selection protocol and representations of patients' medical resource-use are discussed in detail in Section~\ref{data}. All vector representations of patients include age (in years), an indicator variable for gender and positive entries related to the patient's medical resource-use. A patient vector $\mathbf{p}$ can be written in the following general form, where $\dmeds$ and $\dprovs$ denote the number of features in the vectorization of medication and provision use, respectively:

\begin{equation}
\mathbf{p} \in \mathbb{R}^{2 + \dmeds + \dprovs}_+ = 
\begin{bmatrix} 
age 		& gender 	& medication		& provisions  \\
\mathbb{R}_+	& \{0, 1\}	& \mathbb{R}^{\dmeds}_+	& \mathbb{R}^{\dprovs}_+ 
\end{bmatrix}.
\end{equation}
In Sections~\ref{ATC-vectorization} and~\ref{nomenclature} we explain how records related to medication purchases and provisions were represented in vector form. All entries in the vector representations were consistently normalized to the interval $[0, 1]$ by dividing feature-wise by the $99^{th}$ percentile and subsequently clipping where necessary. These normalized vector representations are used as inputs for the learning algorithms described in Section~\ref{learning-methods}.

\begin{figure}[!h]
  \centering
  \includegraphics[width=\textwidth]{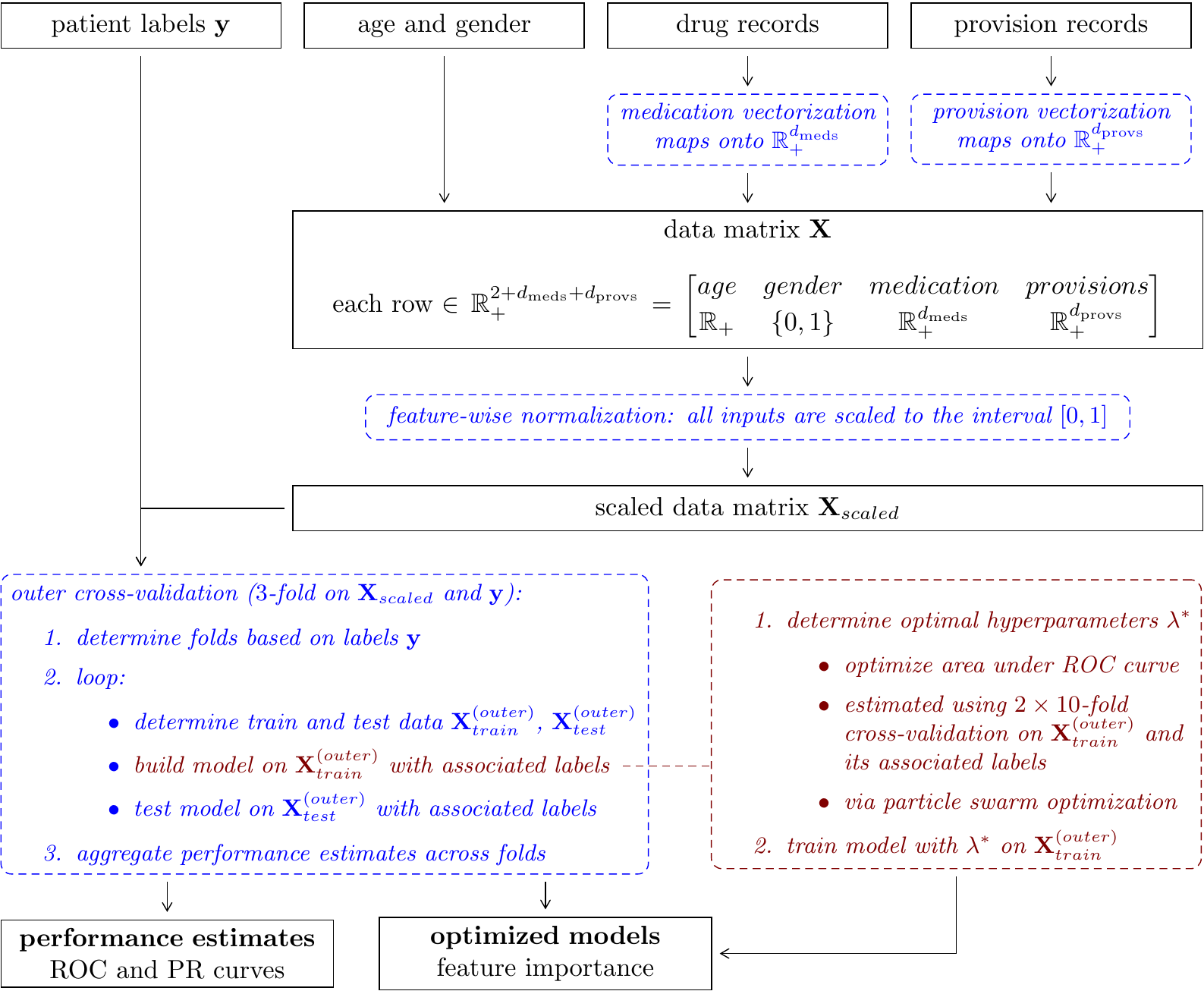}
  \caption{Overview of the full learning approach: data set vectorization, normalization and the nested cross-validation setup. Per iteration, hyperparameter optimization and model training is done based exclusively on $\mathbf{X}_{train}^{(outer)}$.} 
  \label{fig:flowchart}
\end{figure}

Figure~\ref{fig:flowchart} summarizes the full machine learning pipeline, which starts from expenditure records and ends with models to predict whether a patient will start glucose-lowering pharmacotherapy along with an estimate of their generalization performance. We used nested cross-validation to estimate generalization performance of different learning configurations \citep{varma2006bias}. The outer 3-fold cross-validation is used to estimate generalization performance of the full learning approach. Internally, twice iterated 10-fold cross-validation was used to find optimal hyperparameters for every learning method.

\paragraph{Hyperparameter search} 
We used Optunity's particle swarm optimizer to identify suitable hyperparameters for each approach based on the given training set as defined by the outer cross-validation procedure \citep{claesen2014easy}. Every tuple of hyperparameters was evaluated using twice iterated 10-fold cross-validation on the training set. Per technique, the hyperparameters that maximized cross-validated performance were selected and used to train a model on the full training set.

\paragraph{Model evaluation}
Models are compared based on area under the ROC curve. ROC curves visualize a classifier's performance spectrum by depicting its true positive rate (TPR)\footnote{TPR measures the fraction of true positives that are correctly identified by the classifier.} as a function of its false positive rate (FPR)\footnote{FPR measures the fraction of true negatives that are incorrectly identified by the classifier.} while varying the decision threshold to decide on positives. Area under the ROC curve (AUROC) is a useful summary statistic of a classifier's performance. AUROC is equal to the probability that the classifier ranks a random positive higher than a random negative and is known to be equivalent to the Wilcoxon test of ranks \citep{hanley1982meaning}.

\paragraph{Computing ROC curves} Full label knowledge is required to compute ROC curves. In previous work, we introduced a method to compute bounds on ROC curves based on positive and unlabeled data \citep{claesen2015icml}. Briefly, it is based on the positions of known positives in a ranking produced by a given classifier and requires two things:
\begin{itemize}
\item The rank distributions of labeled and latent positives must be comparable. This holds when known and latent positives follow the same distribution in input space (ie. the vector representation of patients). This is a fair assumption in our application, since we specifically ignore records after the start of glucose-lowering pharmacotherapy while identifying the set of positives (see Section~\ref{data}), so the medication regimen of known positives has not yet diverged from the regimen of untreated patients.
\item An estimate $\hat{\beta}$ of the fraction of latent positives in the unlabeled set is needed, that is the fraction of members that have never used GLAs but are likely to start glucose-lowering pharmacotherapy. In the period $2010$--$2014$ roughly $8\%$ of members of NACM aged 40 or higher started using GLAs. Underestimating $\hat{\beta}$ results in an underestimated ROC curve and vice versa  \citep{claesen2015icml}. We opted to be conservative and used $\hat{\beta}_{lo} = 5\%$ to estimate lower bounds and $\hat{\beta}_{up}=10\%$ for upper bounds.
\end{itemize} 
We consistently used the \emph{lower} bounds for hyperparameter search. All our performance reports contain lower and upper bounds, based on $\hat{\beta}_{lo}$ and $\hat{\beta}_{up}$, respectively.

\paragraph{Diagnosing overfitting} 
In addition to measuring performance, we diagnosed overfitting via the concept of rank distributions as defined by \citet{claesen2015icml}. The rank distribution of a subset of test instances is defined as the distribution of the positions of these test instances in a ranking of the full test set based on a model's predicted decision values. We diagnose overfitting based on the rank distributions of known positive training instances ($\mathcal{P}_{train}$) and known positives in the independent test fold ($\mathcal{P}_{test}$) after predicting the full data set. If the model overfits, the rank distribution of $\mathcal{P}_{train}$ is inconsistent with the rank distribution of $\mathcal{P}_{test}$. Specifically, ranks in $\mathcal{P}_{test}$ are worse than those in $\mathcal{P}_{train}$ when the model overfits. This can be quantified via the Mann-Whitney U test \citep{mann1947test} based on ranks of $\mathcal{P}_{train}$ and $\mathcal{P}_{test}$ after predicting the full data set (that is all outer folds). The Mann-Whitney U test is expected to yield a non-significant result when the rank distributions of $\mathcal{P}_{train}$ and $\mathcal{P}_{test}$ are comparable. We report the average $p$-values of the test across outer cross-validation folds for each model (low $p$-values indicate overfitting).


\subsection{Data Set Construction} \label{data}
We constructed a data set containing records of patients born before 1973 (e.g. $40$ or more years old in 2012). Patients with records of glucose-lowering agents (GLAs) during less than 30 days were discarded. Patients with records of glucose-lowering therapy prior to 2012 were discarded. Patients that joined NACM after 2005 were also discarded, as we cannot determine whether these patients used GLAs in the recent past. 

All patients that started glucose-lowering pharmacotherapy in 2012 or later are included as known positives ($n=31,066$), along with unlabeled patients that were sampled at random from the remaining NACM members ($n=79,243$). Known positives have a minimum of 30 days between the first and last purchase of GLAs to avoid contaminating the data set with false positives, for instance due to insulin use in surgical and medical ICUs \citep{van2001intensive, van2006intensive}. It must be noted that some false positives remain, that is patients that use GLAs but not for glycemic control.

In Sections~\ref{ATC-vectorization} and \ref{nomenclature} we describe the vector representations of records regarding medication and medical provisions, respectively.

\subsubsection{Representation of medication records} \label{ATC-vectorization}
The simplest way to represent medication purchases during a time interval is by having one input dimension per active substance (level 5 ATC codes) and counting the purchased volume in terms of DDDs. This representation is easy to construct but fails to capture any similarity between active substances, such as the system or organ on which they act.

\begin{figure}[!h]
  \centering
  \includegraphics[width=\textwidth]{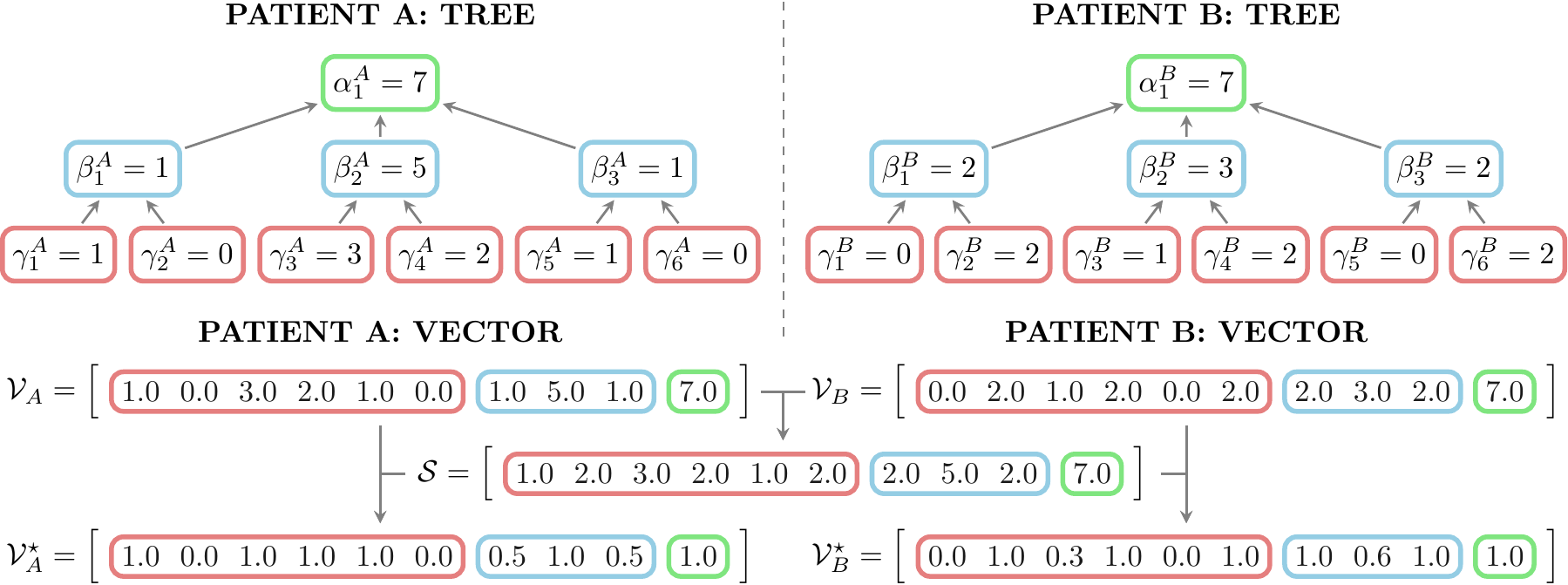}
  \caption{Visualization and vectorization of trees. In the tree representation, the value of internal nodes is the sum of the values of its children. The unnormalized vector representations $\mathcal{V}_A$ and $\mathcal{V}_B$ contain the values per node in the tree representation in some fixed order. Inner products between unnormalized representations $\mathcal{V}_A$ and $\mathcal{V}_B$ are mainly influenced by the top level nodes, since those have the largest value by construction. This undesirable effect can be fixed through feature-wise scaling. The scaling vector $\mathcal{S}$ was constructed using node-wise maxima. The normalized vector representations $\mathcal{V}_A^\star$ and $\mathcal{V}_B^\star$ are obtained by dividing the vector representations ($\mathcal{V}_A$, $\mathcal{V}_B$) element-wise by entries in the scaling vector $\mathcal{S}$. $\mathcal{V}_A^\star$ and $\mathcal{V}_B^\star$ are used as input to classifiers in the remainder of this work. As desired, the inner product of normalized vector representations is increasingly influenced by similarities at higher depths in the tree representations.} 
  \label{fig:medication-vectorization}
\end{figure}

\paragraph{Imposing structure}
We can directly use the hierarchical structure of the ATC system to define a measure of similarity between drugs. To impose structure between drugs we included input dimensions related to more generic levels of the ATC hierarchy (levels 1 to 4). On more generic levels we summed all DDD counts of active substances per category (level 5). This redundancy allowed us to express similarity between different active substances with a standard inner product. By normalizing every feature to the unit interval, we obtained the desired effect that patients with comparable drug use on ATC level 5 are more similar than patients that only share coefficients on more generic levels.
Figure~\ref{fig:medication-vectorization} illustrates this vector representation of trees and the effect of normalization.

\paragraph{Summary} All vectorizations related to drug purchases are described in Table~\ref{table:meds-vects}.

\begin{table}[!h]
\centering
\begin{tabular}{llr}
\toprule
vectorization	 & description	 & $\dmeds$ \\
\midrule
\textsc{atc 5}		 & counts of DDDs per medication class in ATC level 5		& 4,580 \\
\textsc{atc 1--4}	 & counts of DDDs per medication class in ATC levels 1--4 	& 1,257 \\
\textsc{atc 1--5}	 & counts of DDDs per medication class in ATC levels 1--5 	& 5,837 \\
\bottomrule
\end{tabular}
\caption{Summary of vectorization schemes used for records of drug purchases.}
\label{table:meds-vects}
\end{table}

\subsubsection{Representation of provision records} \label{nomenclature}
\begin{figure}[!h]
  \centering
  \includegraphics[width=\textwidth]{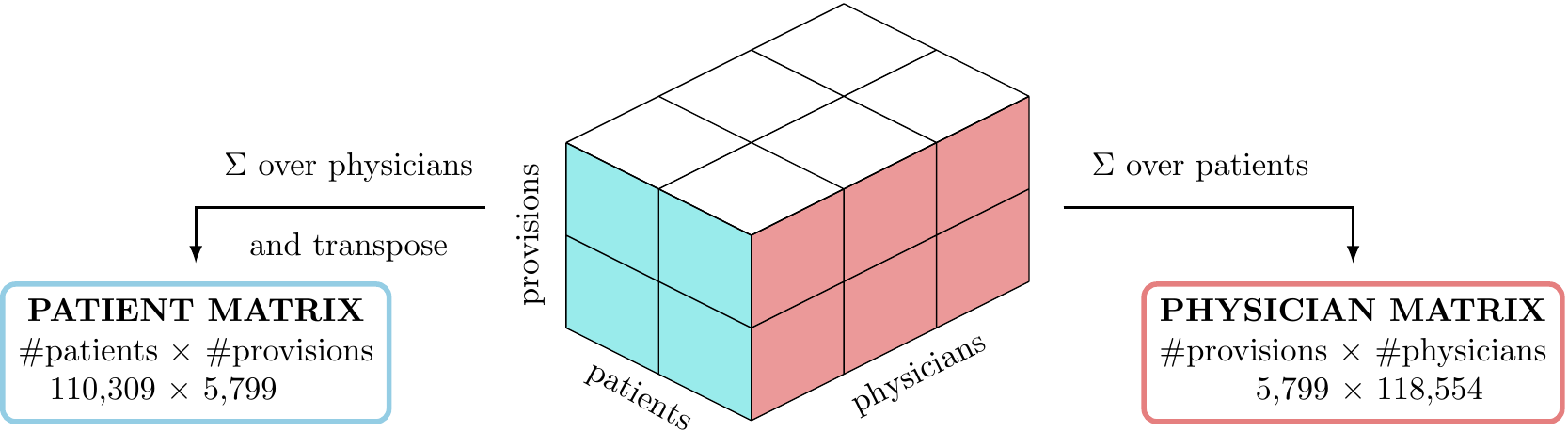}
  \caption{Tensor formulation of medical provisions with three components: patients, physicians and provisions. Each entry in the tensor is the frequency of the given tuple. This provision tensor is very sparse. The patient matrix is obtained by summing counts over all physicians (transposed). The physician matrix is obtained by summing counts over all patients. These matrices capture complementary information.} 
  \label{fig:provisions-tensor}
\end{figure}

When considering a specific time period, we can describe records by a (sparse) three-dimensional tensor containing frequency counts as illustrated in Figure \ref{fig:provisions-tensor}. We filtered all provisions with a description containing \emph{diabetes}, \emph{insulin} and \emph{glucose} and provisions not recorded with a physician identifier. After filtering, 5,799 distinct provision codes remain (denoted by $\# provisions$).

Each patient is modelled by a histogram of their provisions in the period of interest. This essentially means we compute the sum over the $physician$-component of the tensor representation to obtain a matrix, in which rows and columns represent patients and provisions, respectively. Unfortunately, the encoding of provisions has no medically relevant structure in contrast to the ATC hierarchy for drugs as discussed in Section~\ref{ATC-vectorization}.

\begin{figure}[!h]
  \centering
  \includegraphics[width=0.5\textwidth]{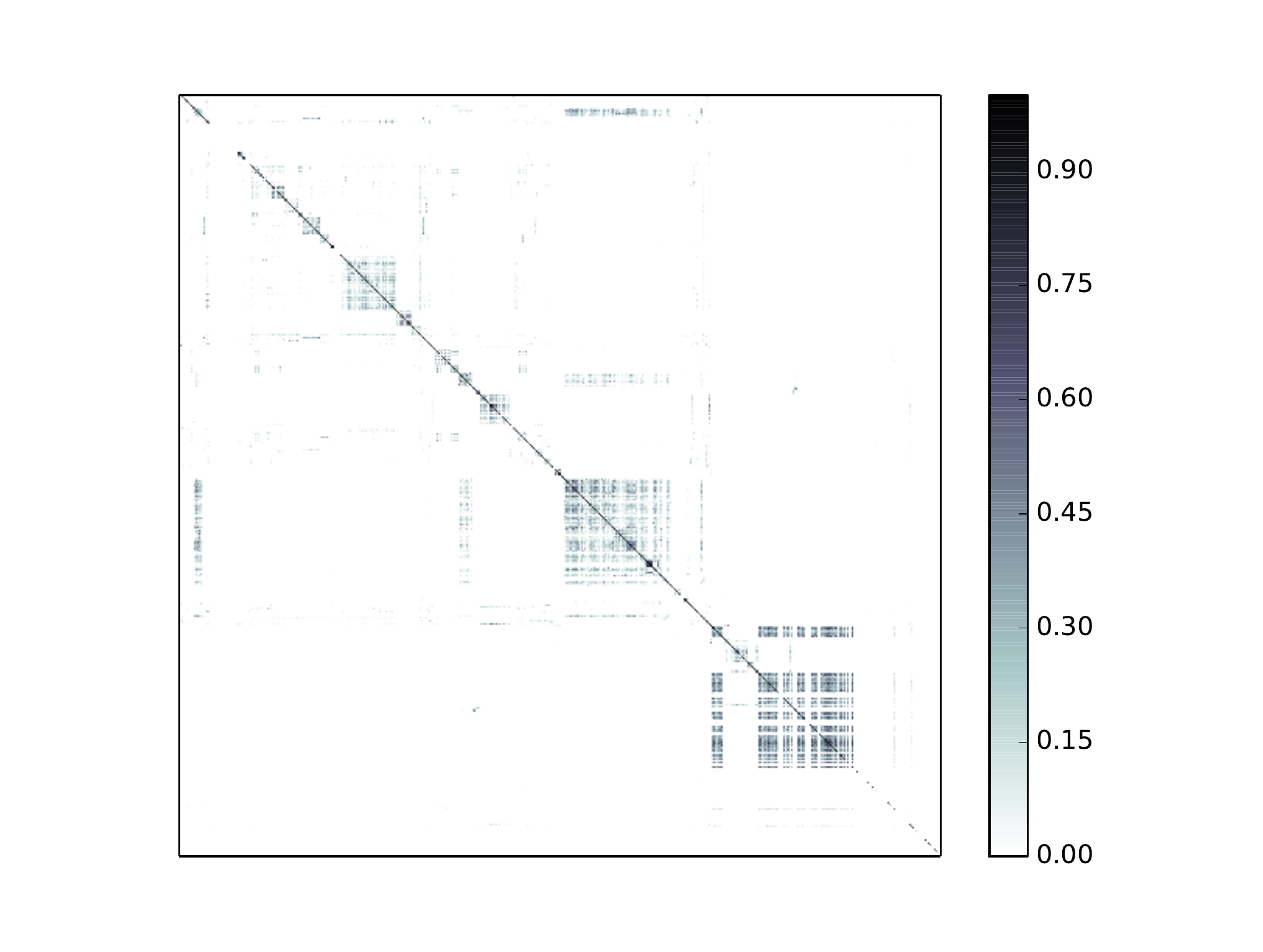}
  \caption{Structure of the provision similarity matrix $\mathbf{S}_{prov}$ based on providing physicians.} 
  \label{fig:Sprovs}
\end{figure}

\paragraph{Imposing structure} In order to define a reasonable similarity measure between patients, we first had to impose a structure onto the nomenclature that captures similarity between provisions. To structure provisions, we should not use information originating from the patient matrix, as this may cause information leaks (since the patient matrix is used directly in our models for prediction). Instead, we used the complementary physician matrix as a basis to define similarity between provisions, which essentially serves as a proxy for the medical specializations to which each provision belongs. We started from cosine similarity between nomenclature codes based on the physician matrix. We used cosine similarity because it is known to work well for text mining with bag-of-words representations, which is comparable to our use case as it also features sparse, high dimensional input spaces. The cosine similarity $\kappa_{cos}$ between two row vectors $\mathbf{u}$ and $\mathbf{v}$ is defined as:
\begin{equation}
\kappa_{cos}(\mathbf{u}, \mathbf{v}) = \frac{\langle \mathbf{u}, \mathbf{v} \rangle}{\|\mathbf{u}\| \cdot \|\mathbf{v}\|} = \frac{\mathbf{u}\mathbf{v}^T}{\|\mathbf{u}\| \cdot \|\mathbf{v}\|}.
\label{eq:cosine}
\end{equation}
Using cosine similarity we can construct a pair-wise similarity matrix $\mathbf{S}_{prov}$ between provisions based on the rows of the physician matrix $\mathbf{x}_i, i=1..\# provisions$:
\begin{equation}
\mathbf{S}_{prov} = \big(\kappa_{cos}(\mathbf{x}_i,\mathbf{x}_j)\big)_{ij} \in \mathbb{R}^{\# provisions \times \# provisions}.
\label{eq:cosine-kernel}
\end{equation}
$\mathbf{S}_{prov}$ expresses similarity between provision codes based on the physicians that provide them and can be regarded as a proxy for the medical subdomain each provision frequently occurs in. In our context, its entries range from $0$ (completely orthogonal) to $+1$ (exact similarity). To impose sparsity we set all entries of $\mathbf{S}_{prov}$ below $0.05$ to $0$. Its structure is visualized in Figure~\ref{fig:Sprovs}, which clearly indicates that our approach successfully identifies some coherent groups of provisions.

Finally, the structured representation of provisions $\mathbf{P}_{struct}$ is defined as the matrix product between the patient matrix $\mathbf{P}_{flat}$ and the provision similarity matrix $\mathbf{S}_{prov}$:
\begin{equation}
\mathbf{P}_{struct} = \mathbf{P}_{flat} \times \mathbf{S}_{prov} \in \mathbb{R}^{\# patients \times \# provisions}.
\end{equation}
$\mathbf{P}_{struct}$ approximately captures which provisions occur in a patient's history with redundancy based on medical specializations.

\paragraph{Summary} All vectorizations related to medical provisions are described in Table~\ref{table:prov-vects}.
\begin{table}[!h]
\centering
\begin{tabular}{lllr}
\toprule
vectorization	 & symbol & description	 & $\dprovs$ \\
\midrule
\textsc{provs flat}	 & $\mathbf{P}_{flat}$ & entries taken from the patient matrix	& $5,799$ \\
\textsc{provs struct}	 & $\mathbf{P}_{struct}$ & captures similarity between provisions	& $5,799$ \\
\textsc{provs both}	 & $\mathbf{P}_{flat}\ |\ \mathbf{P}_{struct}$ & concatenation of flat \& structured  			& $11,598$ \\
\bottomrule
\end{tabular}
\caption{Summary of vectorization schemes used for records of medical provisions.}
\label{table:prov-vects}
\end{table}


\subsection{Modeling approaches} \label{learning-methods}
Having only positive and unlabeled data (PU learning) presents additional challenges for learning algorithms. Two broad classes of approaches exist to tackle these problems: (i) two-phase methods that first attempt to identify likely negatives from the unlabeled set and then train a supervised model on the positives and inferred negatives \citep{liu02partially, Yu:2005:SCM:1108759.1108762} and (ii) approaches that treat the unlabeled set as negatives with label noise \citep{Elkan:2008:LCO:1401890.1401920,Lee03learningwith,MORDELET-2010-523336,Claesen2015resvm}. 

We have tested three approaches from the latter category in this work, namely class-weighted SVM \citep{Liu:2003:BTC:951949.952139}, bagging SVM \citep{MORDELET-2010-523336} and the robust ensemble of SVM models \citep{Claesen2015resvm}. All of these approaches are based on support vector machines. We used the linear kernel on vector representations of patients as described in Section~\ref{data}.\footnote{Though it must be noted that the ensemble methods are always implicitly nonlinear.} We will briefly introduce each method in the following subsections.

\subsubsection{Class-weighted SVM} \label{bsvm}
Class-weighted SVM (CWSVM) uses a misclassification penalty per class. CWSVM was first applied in a PU learning context by \citet{Liu:2003:BTC:951949.952139}, by considering the unlabeled set to be negative with noise on its labels. A CWSVM is trained to distinguish positives ($\mathcal{P}$) from unlabeled instances ($\mathcal{U}$), leading to the following optimization problem:
\begin{align}
\min_{\alpha,\xi,b}\ & \frac{1}{2}\sum_{i=1}^N\sum_{j=1}^N \alpha_i\alpha_j y_i y_j \kappa(\mathbf{x}_i,\mathbf{x}_j)+C_{\mathcal{P}}\sum_{i \in\mathcal{P}} \xi_i + C_{\mathcal{U}}\sum_{i\in\mathcal{U}} \xi_i, \label{eq:bsvm} \\
\text{s.t. } &y_i(\sum_{j=1}^N \alpha_j y_j \kappa(\mathbf{x}_i,\mathbf{x}_j)+b)\geq 1-\xi_i, &i=1,\ldots,N, \nonumber \\
&\xi_i \geq 0, &i=1,\ldots,N, \nonumber
\end{align}
where $\alpha \in \mathbb{R}^N$ are the support values, $\mathbf{y} \in \{-1,+1\}^N$ is the label vector, $\kappa(\cdot,\cdot)$ is the kernel function, $b$ is the bias term and $\xi \in \mathbb{R}^N$ are the slack variables for soft-margin classification. The misclassification penalties $C_{\mathcal{P}}$ and $C_{\mathcal{U}}$ require tuning. We used the implementation available in scikit-learn \citep{pedregosa2011scikit} based on LIBSVM \citep{CC01a}.

\subsubsection{Bagging SVM} \label{baggingsvm}
In bagging SVM, random resamples are drawn from the unlabeled set and CWSVM classifiers are trained to discriminate all positives from each resample \citep{MORDELET-2010-523336}. Resampling the unlabeled set induces variability in the base models which is exploited via bagging. Base model predictions are aggregated via majority voting.

Bagging SVM with linear base models has two hyperparameters, namely the size of resamples of the unlabeled set $n_\mathcal{U}$ and the misclassification penalty on unlabeled instances $C_\mathcal{U}$. The misclassification penalty on positives $C_\mathcal{P}$ is fixed via the following rule:
\begin{equation}
C_{\mathcal{P}} = \frac{n_{\mathcal{U}} \times C_{\mathcal{U}}}{|\mathcal{P}|}, \label{eq:bagpenalties}
\end{equation}
where $|\mathcal{P}|$ denotes the number of known positives. The heuristic rule in Equation~\ref{eq:bagpenalties} is common in imbalanced settings \citep{cawley2006leave,daemen2009kernel}. We implemented bagging SVM using the EnsembleSVM library \citep{JMLR:v15:claesen14a}.

\subsubsection{Robust ensemble of SVM models}
The robust ensemble of SVM models (RESVM) is a modified version of bagging SVM in which both the positive and unlabeled sets are resampled when constructing base model training sets \citep{Claesen2015resvm}. The extra resampling induces additional variability between base models which improves performance when combined with a majority vote aggregation scheme. \citet{Claesen2015resvm} demonstrated that resampling the positive set provides robustness against false positives, which makes RESVM appealing for our application since our data set is known to contain a small fraction of false positives (as explained in Section~\ref{data}).

When using linear base models, the RESVM approach has four hyperparameters that must be tuned, namely resample sizes and misclassification penalties per class. This approach was implemented based on EnsembleSVM \citep{JMLR:v15:claesen14a}.

%
%

\section{Results and discussion}
Section~\ref{benchmark} shows the predictive performance per learning configuration and compares these performances to the current state-of-the-art in large-scale risk assessment for T2D. Section~\ref{resvmroc} shows performance curves of the best configuration, which enable us to determine suitable cutoffs to identify target groups in practice. Finally, Section~\ref{features} describes a simple approach to assess which features contribute most to risk according to our best models.

\subsection{Benchmark of learning methods} \label{benchmark}
Table~\ref{table:results} summarizes the performance of each learning configuration. The \textsc{age,gender} feature set provides a baseline for comparison, all other feature sets include these as well. As shown in the results, this two-dimensional representation already carries some information.

\begin{table}[!h]
\centering
\begin{tabular}{lcccccccc}
\toprule
 		 & \multicolumn{2}{c}{RESVM} 	 & & \multicolumn{2}{c}{bagging SVM} 	& & \multicolumn{2}{c}{class-weighted SVM} \\ \cline{2-3} \cline{5-6} \cline{8-9}
features	 & AUROC ($\%$)	 & $p$ 		 & & AUROC ($\%$) & $p$			& & AUROC ($\%$) & $p$ \\
\midrule
\textsc{age, gender}	 & $55.74$--$56.64$ & $*$ & & $58.61$--$59.67$ & $*$ & & $\mathbf{60.96}$--$\mathbf{62.21}$ & $0.04$ \\
\textsc{atc 5}		 & $\mathbf{72.55}$--$\mathbf{74.43}$ 	& $0.17$ & & $70.83$--$72.62$ & $0.09$ & & $71.89$--$73.74$ 			& $0.01$ \\
\textsc{atc 1--4}	 & $\mathbf{73.12}$--$\mathbf{75.07}$ & $0.07$ & & $69.57$--$71.24$ & $*$ & & $73.05$--$74.91$ & $0.04$ \\
\textsc{atc 1--5} 	 & $\mathbf{74.34}$--$\mathbf{76.27}$ 	& $0.13$ & & $71.50$--$73.27$ & $0.05$ & & $72.13$--$73.94$ 			& $*$ \\
\textsc{provs flat}	 & $58.45$--$59.51$ & $*$ & & $60.74$--$61.92$ & $*$ & & $\mathbf{63.01}$--$\mathbf{64.31}$ & $*$ \\
\textsc{provs struct}	 & $57.40$--$58.39$ & $0.02$ & & $59.53$--$60.58$ & $0.01$ & & $\mathbf{62.53}$--$\mathbf{63.81}$ & $0.01$ \\
\textsc{provs both}	 & $58.89$--$59.75$ & $*$ & & $61.72$--$62.87$ & $*$ & & $\mathbf{63.45}$--$\mathbf{64.75}$ & $*$ \\
\textsc{atc $|$ provs}	 & $\mathbf{74.89}$--$\mathbf{76.82}$ & $0.04$ & & $69.72$--$71.40$ & $*$ & & $73.77$--$75.64$ & $*$ \\

\bottomrule
\end{tabular}
\caption{Average bounds on area under the ROC curve and $p$-value of the Mann-Whitney U test over all folds for different feature sets per learning approach in a long-term prediction setup. The lower and upper bounds on AUC were computed with $\hat{\beta}_{lo}=0.05$ and $\hat{\beta}_{up}=0.10$, respectively. The \mbox{\textsc{atc $|$ provs}} feature set is the concatenation of the best performing sets per aspect, namely \textsc{atc 1--5} and \textsc{provs both}. Stars ($*$) denote $p$-values below $0.005$.
}
\label{table:results}
\end{table}

Based on Table~\ref{table:results} we can conclude that a patient's medication history is highly informative to predict the start of GLA therapy. Using features based on ATC level 5, the RESVM model obtained an AUC between $72.55\%$ and $74.43\%$. By adding redundancy as described in Section~\ref{ATC-vectorization} the performance based on medication history alone was further increased to between $74.34\%$ and $76.27\%$ for the best learning approach (RESVM).

Predictive performance based on provisions alone turned out fairly poor, showing only a mild improvement compared to models based exclusively on age and gender for all learning algorithms. Interestingly, the best approach for representations based on provisions was class-weighted SVM, with RESVM being worst of all three learning methods. It appears that for these representations, large training sets are better: class-weighted SVM uses the full training set, bagging SVM uses all positives and a subset of unlabeleed instances per base model and RESVM uses (small) subsets of both positives and unlabeled instances per base model.

The best representation included age, gender, and structured information about drugs and provision history of the patient. The best learning method on this representation was RESVM, achieving an AUC between $74.89\%$ and $76.82\%$. In Section~\ref{stateoftheart} we compare the performance of our approach to competing screening methods.

Finally, RESVM appears most resistant to overfitting in the hyperparameter optimization stage as it consistently exhibits the highest average $p$-values in our diagnostic test (higher is better, see Section~\ref{setup}). We believe this to be attributable to the use of small resamples of both positives and unlabeled instances when training base models in RESVM, since this makes it unlikely to obtain a structural overfit of the ensemble model on the full training set. In contrast, bagging SVM is far more prone to overfitting because every base model is trained on all positives.

\subsubsection{Comparison to state-of-the-art} \label{stateoftheart}
Our best approach obtained cross-validated AUC between $74.89\%$ and $76.82\%$ (exact numbers are unknown due to the lack of known negatives). This is comparable to many competing approaches, based on questionnaires and some clinical information such as the Cambridge Risk Score (AUC 67\%--80\%, \citep{spijkerman2004performance, griffin2000diabetes}), the Danish risk score (AUC $72\%$--$87.6\%$, \citep{glumer2004danish}), the German diabetes risk score (AUC $75\%$--$83\%$, \citep{schulze2007accurate}) and a Dutch approach (AUC 74\%, \citep{baan1999performance}). Approaches using detailed clinical information generally perform better, but are more expensive to maintain \citep{stern2002identification, mcneely2003comparison, lindstrom2003diabetes, heikes2008diabetes}.
 They key advantage of our approach is the fact it is easy to implement on a population wide scale at virtually no operational cost.

The target class we used in this work is stricter than in the risk prediction methods mentioned in Section~\ref{sec:stateoftheart}, namely patients that require GLAs for glycemic control versus patients with impaired glycemic control, respectively (except for \citet{lindstrom2003diabetes}, which also predicted drug-treated T2D). It is reasonable to assume that our models generally rank patients with impaired glycemic control but without a need for GLAs higher than patients without impaired glycemic control. In our performance assessment both of these patient groups are essentially treated as negatives, in contrast to the screening programmes mentioned previously which treat patients with impaired glycemic control as positives. Hence, we believe the performance of our models would appear higher when evaluated against a target class comprising all patients with impaired glycemic control, as is done in the evaluation of other screening approaches. Unfortunately, we are unable to accurately identify patients with impaired glycemic control but without need for GLAs.

All competing methods use either clinical information or direct knowledge of risk factors that is unavailable to us. Furthermore, the characteristics that are lacking in our data have been reported to be the most informative to assess risk for T2D \citep{lindstrom2003diabetes,stern2002identification, mcneely2003comparison}. We obtained generalization performances that are comparable to existing approaches, despite these missing predictors. Finally, our approach is the only one that is based exclusively on existing data that is always available, without requiring additional patient contacts or clinical tests.

\subsection{Receiver Operating Characteristic curves for RESVM} \label{resvmroc}
\begin{figure}[!h]
\centering
\subfigure[Receiver Operating Characteristic curve.]{\includegraphics[width=0.45\textwidth]{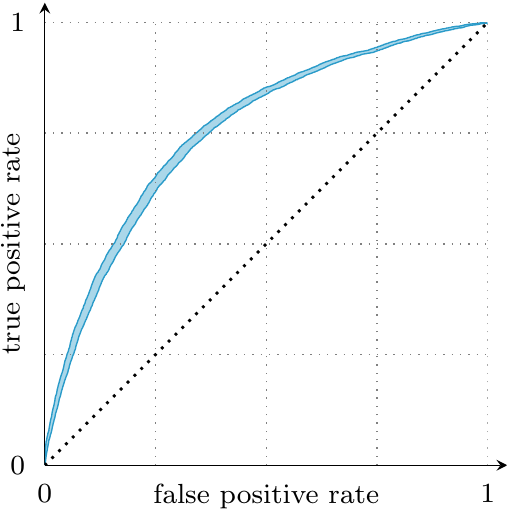}\label{fig:roc}}\qquad
\subfigure[Precision-Recall curve.]{\includegraphics[width=0.45\textwidth]{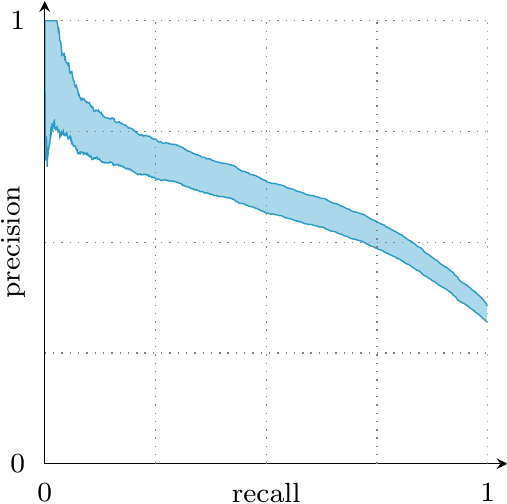}\label{fig:pr}}\\
\caption{Performance curves for the best model: RESVM classifier based on \textsc{atc $|$ provs} vectorization. The lower and upper bounds are estimated using $\hat{\beta}_{lo}=5\%$ and $\hat{\beta}_{up}=10\%$, respectively. }
\label{fig:performance-curves}
\end{figure}

The RESVM model based on \textsc{atc $|$ provs} vectorization had the best overall performance. Figure~\ref{fig:performance-curves} shows bounds on the ROC and PR curves for this model. These bounds were computed using the technique described by \citet{claesen2015icml}. The true curve is unknown because we do not dispose of negative labels.

ROC curves enable us to determine a cutoff to use in practice, based on a suitable balance between true and false positive rate (sensitivity and 1-specificity, respectively). Determining a suitable balance requires a tradeoff between the relative importance of identifying undiagnosed patients (true positives) vis-\`a-vis increased amounts of screening tests on patients that are in fact healthy (false positives).

It should be noted that precision depends on class balance, and therefore the PR curve shown in Figure~\ref{fig:pr} is not representative for screening an overall population, since the overall population has a higher fraction of negatives than our custom data set (i.e. precision would be lower in practice). In contrast, the bounds in ROC space are representative because ROC curves are insensitive to changes in class distribution \citep{Fawcett:2006:IRA:1159473.1159475}.

\subsection{Feature importance analysis for the RESVM model} \label{features}
The RESVM model is implicitly nonlinear due to its majority voting rule to aggregate base model decisions, which poses problems in assessing the importance of each predictor. However, our use of linear base models enables a simple approximation. The decision value for base model $i \in \{1,\ldots,n_{\rm models}\}$ for a test instance $\mathbf{z}$ can be written as follows:
\begin{equation*}
f_i(\mathbf{z}) = \langle \mathbf{w}_i, \mathbf{z}\rangle + \rho_i,
\end{equation*}
where $\mathbf{w}_i$ is the separating hyperplane and $\rho_i$ is a bias term. A simple linear approximation of such ensemble models can be computed as the average of all base model hyperplanes:
\begin{equation*}
\bar{\mathbf{w}} = \sum_{i=1}^{n_{\rm models}} \mathbf{w}_i / n_{\rm models}.
\end{equation*}
Feature importance can then be determined based on the coefficients in $\bar{\mathbf{w}}$. Since we normalized all features to the unit interval $[0,1]$ we can conclude that the features with largest (positive) coefficients contribute most to risk as identified by our model. 

Via this approach, the risk associated to use of cardiovascular medication (ATC main category \textsc{c}) far outweights all other ATC main categories. This is not surprising, as diabetes is known to be strongly related to cardiovascular problems \citep{doi:10.1001/jama.1979.03290450033020, grundy1999diabetes, hu2002elevated}. The relative importance of features will be discussed in detail in a subsequent medical paper.

%
%

\section{Conclusion}
In this work we have demonstrated the ability to predict clinical outcomes based solely on readily available health expenditure data. We successfully built proof-of-concept classifiers to predict the start of glucose-lowering pharmacotherapy in patients above 40. Our experiments show that accurate predictions can be made based on historical medication purchases. These predictions can be further improved by incorporating information about medical provisions and the use of appropriate vectorization schemes.

Since adult patients starting glucose-lowering pharmacotherapy are mainly afflicted with type 2 diabetes (T2D), our models can be used for T2D risk assessment. Our approach presents a novel method for case finding which can be easily incorporated in modern healthcare, since all required data is already available. The associated operational costs are very low as the entire workflow can be fully automated without any need for patient contacts or medical tests. As such, our work provides an efficient and cost-effective method to identify a high risk subgroup, which can then be screened using decisive clinical tests.

Interestingly, our approach works well even though health expenditure data contains very limited direct information on some important known risk factors. In that sense, our approach is fundamentally different from the current state-of-the-art which mainly focuses on quantifying known risk factors directly, either by asking the patient or through clinical tests. The performance of our approach is expected to improve further when additional information about these risk factors can be obtained, e.g. family history and lifestyle.

\section*{Acknowledgments}
The authors wish to thank Bernard Debbaut, Frie Niesten, Koen Cornelis and Michiel Callens for their valuable input in various aspects of this study. 

This study was supported by the Flemish Government (FWO: projects:  G.0871.12N (Neural circuits); IWT: TBM-Logic Insulin(100793), TBM Rectal Cancer(100783), TBM IETA(130256); PhD grants; Industrial Research fund (IOF): IOF Fellowship 13-0260; iMinds Medical Information Technologies SBO 2015, ICON projects (MSIpad, MyHealthData); VLK Stichting E. van der Schueren: rectal cancer) and the Belgian Federal Government (FOD: Cancer Plan 2012-2015 KPC-29-023 (prostate); COST: Action: BM1104: Mass Spectrometry Imaging). M.C. is funded by a PhD grant awarded by IWT (\#111065). P.G is funded by a clinical research foundation of the University Hospitals Leuven-KUL.

\bibliography{bibliography}

\end{document}